\documentclass[conference]{IEEEtran}
\IEEEoverridecommandlockouts

\usepackage{cite}
\usepackage{amsmath,amssymb,amsfonts}
\usepackage{algorithm}
\usepackage{algorithmic}
\usepackage{graphicx}
\usepackage{textcomp}
\usepackage{booktabs}
\usepackage{multirow}
\usepackage{siunitx}
\usepackage{xcolor}
\usepackage{tabularx}
\usepackage{threeparttable}
\usepackage{makecell}
\usepackage{pifont}
\usepackage{tcolorbox}
\usepackage{subcaption}
\usepackage[hidelinks]{hyperref}
\usepackage{adjustbox} 

\usepackage{array}
\usepackage{adjustbox}
\usepackage{ragged2e}
\newcolumntype{Y}{>{\RaggedRight\arraybackslash}X} 
\newcolumntype{C}{>{\centering\arraybackslash}X}   

\usepackage[font=footnotesize]{caption}
\def\BibTeX{{\rm B\kern-.05em{\sc i\kern-.025em b}\kern-.08em
T\kern-.1667em\lower.7ex\hbox{E}\kern-.125emX}}

\renewcommand{\arraystretch}{1.1}
\begin{document}

\title{How Small Can You Go? Compact Language Models for On-Device Critical Error Detection in Machine Translation
\thanks{This research has been funded by the Federal Ministry of Education and Research of Germany and the state of North-Rhine Westphalia as part of the Lamarr Institute for Machine Learning and Artificial Intelligence.}}

\author{
\IEEEauthorblockN{Muskaan Chopra\IEEEauthorrefmark{2}\IEEEauthorrefmark{4},
Lorenz Sparrenberg\IEEEauthorrefmark{2}\IEEEauthorrefmark{4},
Sarthak Khanna\IEEEauthorrefmark{2},
Rafet Sifa\IEEEauthorrefmark{9}\IEEEauthorrefmark{2}\IEEEauthorrefmark{4}}
\IEEEauthorblockA{\IEEEauthorrefmark{9}Fraunhofer IAIS - Department of Media Engineering, Germany}
\IEEEauthorblockA{\IEEEauthorrefmark{2}University of Bonn - Department of Computer Science, Germany}
\IEEEauthorblockA{\IEEEauthorrefmark{4}Lamarr Institute for Machine Learning and Artificial Intelligence, Germany}
}

\maketitle

\begin{abstract}
Large Language Models (LLMs) excel at evaluating machine translation (MT), but their scale and cost hinder deployment on edge devices and in privacy-sensitive workflows. We ask: how small can you get while still detecting meaning-altering translation errors? Focusing on English→German Critical Error Detection (CED), we benchmark sub-2 B models (LFM2-350M, Qwen-3 0.6B/1.7B, Llama-3.2-1B-Instruct, Gemma-3-1B) across WMT21, WMT22, and SynCED-EnDe 2025.
Our framework standardizes prompts, applies lightweight logit-bias calibration and majority voting, and reports both semantic quality (MCC, F1-ERR/F1-NOT) and compute metrics (VRAM, latency, throughput).
Results reveal a clear sweet spot around one billion parameters: Gemma-3-1B provides the best quality-efficiency trade-off, reaching MCC = 0.77 with F1-ERR = 0.98 on SynCED-EnDe 2025 after merged-weights fine-tuning, while maintaining 400 ms single-sample latency on a MacBook Pro M4 Pro (24 GB).
At larger scale, Qwen-3-1.7B attains the highest absolute MCC (+0.11 over Gemma) but with higher compute cost. In contrast, ultra-small models ($< \textbf{0.6}$
B) remain usable with few-shot calibration yet under-detect entity and number errors.
Overall, compact, instruction-tuned LLMs-augmented with lightweight calibration and small-sample supervision, can deliver trustworthy, on-device CED for MT, enabling private, low-cost error screening in real-world translation pipelines.
All datasets, prompts, and scripts are publicly available at our GitHub repository.\footnote{The code is available at: \url{https://github.com/muskaan712/How-small-can-you-get}}

\end{abstract}


\section{Introduction}
\label{sec:intro}

\IEEEPARstart{L}{arge} Language Models (LLMs) have transformed natural language processing, enabling major advances in translation, summarisation, dialogue, and reasoning. Yet their benefits remain unevenly distributed across linguistic and computational boundaries. Most state-of-the-art models are trained primarily on English and other high-resource languages, and their enormous scale requires cloud-based infrastructure. As a result, low-resource languages and low-power devices are often excluded from the LLM revolution. This imbalance raises a pressing question for multilingual AI: how can we make language intelligence universally accessible without depending on ever-larger, centralized models?

We address this question through the task of \emph{Critical Error Detection (CED)} in English→German translation, a high-stakes evaluation problem that requires identifying meaning-altering errors such as hallucinations, omissions, or wrong entities. Unlike general quality estimation, which measures overall fluency or adequacy, CED isolates \emph{critical} semantic divergences that could misinform users or compromise safety. It therefore provides a sensitive diagnostic task for testing whether compact LLMs can reason about translation faithfulness.

\subsection{From Big Data to Small Models}

Large multilingual encoders such as mBERT and XLM-R have demonstrated the potential of cross-lingual pre-training at scale, but they remain computationally heavy. At the same time, hardware progress is bringing transformer inference to consumer devices: recent systems like Apple’s AirPods Pro 3 \cite{airpods3} and Meta’s Ray-Ban Meta Gen 2 \cite{metaglasses} integrate neural cores capable of running small LLMs locally. This shift enables privacy-preserving, offline applications but also demands models that can operate efficiently within tight latency and memory budgets.

While frontier-scale models such as GPT-4 \cite{openai2024gpt4} or GPT-4o \cite{openai2024gpt4o} deliver exceptional translation-evaluation performance, their inference cost prevents deployment in constrained environments. This motivates our investigation of \emph{sub-2B-parameter} models for CED. We benchmark and fine-tune compact LLMs, including Qwen 3-0.6B, LFM2-350M, Gemma 3-1B, Llama 3.2-1B-Instruct, and Qwen 3-1.7B, across three representative datasets: \textsc{WMT21}, \textsc{WMT22}, and the curated \textsc{SynCED-EnDe 2025} resource \cite{anonymous}. Together, these sets provide a comprehensive view of model robustness under varied translation conditions.

\subsection{Critical Error Detection as a Safety Benchmark}

CED bridges translation quality estimation \cite{specia2020findings,federmann2021findings,federmann2022findings} with factual-consistency evaluation frameworks such as COMET \cite{rei2020comet} and MQM \cite{lommel2014multidimensional}. It requires models to go beyond lexical overlap and capture deep semantic mismatches. Previous research on contradiction detection in German \cite{9003090,pielka2020contradiction} showed that smaller neural models can capture subtle inconsistencies when properly trained, but these systems lacked generalization to open-domain translation outputs. Modern LLMs, when paired with carefully designed prompts and calibration, can unify these capabilities within a single multilingual reasoning framework.

\subsection{Motivation: Why Small Models Matter}

Compact models are not merely efficient approximations; they enable new possibilities for inclusion, privacy, and sustainability. In professional translation workflows, finance \cite{finance} \cite{cdfr}, legal \cite{legal}, and healthcare \cite{health} confidentiality often preclude sending text to cloud APIs. Deploying lightweight CED evaluators locally preserves data sovereignty while reducing inference latency and energy consumption. Moreover, smaller models can democratise translation evaluation by allowing low-resource institutions to audit outputs on standard hardware.

This study therefore, pursues two goals:  
(1) to empirically quantify the trade-offs between model size, accuracy, and compute efficiency for multilingual CED; and  
(2) to assess whether compact, instruction-tuned models can match the reasoning reliability of large evaluators when augmented with calibration and fine-tuning.  
Our experiments show that 1 to 2 B-parameter models, such as Gemma 3-1B and Qwen 3-1.7B, achieve competitive Matthews correlation and F1-ERR scores while maintaining latency and memory footprints suitable for edge deployment.

\subsection{Contributions}

This paper makes three key contributions:
\begin{itemize}
    \item \textbf{A systematic scaling study} of compact LLMs for multilingual \emph{Critical Error Detection}, evaluating zero-shot, few-shot, majority-voting, and fine-tuned configurations across \textsc{WMT21}, \textsc{WMT22}, and \textsc{SynCED-EnDe}.
    \item \textbf{An open, reproducible framework} providing fine-tuning scripts, calibration prompts, and compute-efficiency benchmarks (memory, latency, throughput).
    \item \textbf{A demonstration of on-device feasibility}, showing that models under 2 B parameters can deliver reliable multilingual error detection with low latency and high interpretability, paving the way for equitable, sustainable, and accessible language technologies.
\end{itemize}

These findings suggest that linguistic inclusivity and safety do not require ever-larger architectures. With thoughtful calibration and targeted fine-tuning, compact models can deliver trustworthy multilingual reasoning, even on the smallest of devices.

\section{Related Work}
\label{sec:related}

\subsection{Machine Translation Evaluation and Quality Estimation}

Automatic machine translation (MT) evaluation has long relied on reference-based similarity metrics such as BLEU \cite{papineni2002bleu} and METEOR \cite{banerjee2005meteor}.  
While computationally efficient, these n-gram overlap scores correlate only weakly with human judgments of meaning preservation.  
Subsequent learned metrics such as COMET \cite{rei2020comet} improved this alignment by leveraging multilingual encoders and regression heads trained on human-rated examples.  
Complementary human frameworks such as MQM \cite{lommel2014multidimensional} introduced structured taxonomies of translation errors, forming the basis of evaluation protocols in WMT shared tasks \cite{specia2020findings,federmann2021findings,federmann2022findings}.  
However, most existing automatic metrics emphasise average quality rather than identifying \emph{critical} errors that distort factual or logical meaning.  

Recent studies explored whether LLMs can act as evaluators by reasoning directly about semantic fidelity.  
Kocmi and Federmann \cite{kocmi2023lm_eval} demonstrated that GPT-4 achieves state-of-the-art correlations with MQM scores, while Lu et al.\ \cite{lu2023erroranalysis} introduced \emph{error-analysis prompting} to obtain human-like rationales from LLMs.  
Peng et al.\ \cite{peng2023towards} further showed that ChatGPT can perform comparative translation ranking using carefully tuned instructions.  
Although these works highlight the potential of large proprietary models, they remain limited by size, cost, and reproducibility, motivating research into compact, open, and transparent evaluators.

\subsection{Critical Error Detection and Translation Safety}

Critical Error Detection (CED) narrows the focus of evaluation to meaning-altering mistakes such as hallucinations, entity mismatches, or negation flips.  
It connects to earlier work on factual inconsistency and contradiction detection in German \cite{9003090,pielka2020contradiction}, which showed that neural classifiers can capture semantic polarity differences even without reference translations.  
More recently, Pucknat et al.\ \cite{pucknat2022informed} proposed informed pre-training objectives for English-German CED, while Jung et al.\ \cite{jung2024explainable} released the \textsc{Explainable CED} dataset with human-annotated rationales.  
The 2025 \textsc{SynCED-EnDe} resource \cite{anonymous} extended this direction through balanced, human-validated annotations of \texttt{ERR}/\texttt{NOT} labels, providing a reproducible testbed for multilingual error detection.  

CED thus serves as a realistic proxy for translation-safety monitoring, complementing industrial studies such as Pielka et al.\ \cite{pielka2025translation}, who applied LLMs to detect translation anomalies in financial documents.  
Our work builds on these efforts by systematically examining whether small-scale, instruction-tuned LLMs can perform CED reasoning with comparable reliability to frontier-scale evaluators.

\subsection{Efficient Fine-Tuning and Compact LLMs}

Scaling down LLMs while preserving reasoning ability has become a central research challenge.  
Parameter-efficient fine-tuning techniques such as LoRA and PEFT have reduced training cost by adapting a small subset of weights, while quantisation and distillation enable deployment on limited hardware.  
Toolkits like \emph{Unsloth} \cite{unsloth2024} streamline such workflows, and recent architectures, including ModernBERT \cite{modernbert} and mmBERT \cite{mmbert}, demonstrate that careful tokenization, attention sparsity, and language-adaptive scheduling can yield faster and memory-efficient multilingual encoders.  

These advances intersect with a broader movement toward open, transparent model families.  
Meta’s LLaMA 3 and 3.3 releases \cite{meta2024llama3,meta2025llama33} and OpenAI-OSS reproductions \cite{gptoss2025report,gptoss2025lora} provide accessible backbones for experimentation under permissive licences.  
Despite their availability, systematic benchmarks of such small-parameter models on safety-critical multilingual tasks remain rare.  
Our study contributes the first comparative evaluation of models below 2 B parameters, Gemma 3-1B, Qwen 3-1.7B, and others on the CED task, uniting perspectives from translation evaluation, factual consistency, and efficient model adaptation.

\section{Task Definition and Datasets}
\label{sec:task}

\subsection{Critical Error Detection (CED)}
Given an English source sentence \(S\) and a German translation \(T\), the goal is binary classification:
\[
\mathrm{CED}(S,T) \in \{\texttt{ERR}, \texttt{NOT}\},
\]
where \texttt{ERR} denotes a meaning-altering error (e.g., omission, hallucination, entity/number mismatch, negation flip, or safety-critical distortion) and \texttt{NOT} denotes preserved meaning. 
We evaluate three inference regimes: (i) zero-shot, (ii) few-shot with optional majority voting, and (iii) fine-tuned (merged full weights). 
Metrics include accuracy, F1 for the \texttt{ERR} class, and Matthews Correlation Coefficient (MCC).

\subsection{Dataset Statistics (Sentence-level (S,T) pairs)}
All counts refer to \emph{sentence pairs} \((S_{\text{en}}, T_{\text{de}})\) with a single CED label per pair.

\begin{table}[t]
  \centering
  \caption{Dataset splits by number of sentence pairs.}
  \label{tab:data_splits}
  \begin{threeparttable}
    \begin{adjustbox}{max width=\columnwidth}
      {\scriptsize
      \setlength{\tabcolsep}{6pt}        
      \renewcommand{\arraystretch}{1.1}  
      \begin{tabular}{lcc}
        \toprule
        \textbf{Dataset} & \textbf{Train \#pairs} & \textbf{Dev \#pairs} \\
        \midrule
        WMT21             & 8{,}000   & 1{,}000 \\
        WMT22             & 155{,}511 & 17{,}280 \\
        SynCED-EnDe 2025 & 8{,}000   & 1{,}000 \\
        \bottomrule
      \end{tabular}
      }
    \end{adjustbox}
  \end{threeparttable}
\end{table}

\begin{table}[t]
  \centering
  \caption{Label distributions per split. For WMT22, OK $\rightarrow$ NOT and BAD $\rightarrow$ ERR.}
  \label{tab:label_dist}
  \begin{threeparttable}
    \begin{adjustbox}{max width=\columnwidth}
      {\scriptsize
      \setlength{\tabcolsep}{6pt}
      \renewcommand{\arraystretch}{1.1}
      \begin{tabular}{lcccc}
        \toprule
        \multirow{2}{*}{\textbf{Dataset}} &
        \multicolumn{2}{c}{\textbf{Train}} &
        \multicolumn{2}{c}{\textbf{Dev}} \\
        \cmidrule(lr){2-3}\cmidrule(lr){4-5}
        & NOT & ERR & NOT & ERR \\
        \midrule
        WMT21              & 5{,}760 & 2{,}240 & 700  & 300 \\
        WMT22 (OK/BAD)     & 146{,}574 & 8{,}937 & 16{,}329 & 951 \\
        SynCED-EnDe 2025  & 4{,}000 & 4{,}000 & 500  & 500 \\
        \bottomrule
      \end{tabular}
      }
    \end{adjustbox}
  \end{threeparttable}
\end{table}

\subsection{Model Families Evaluated}
We target compact models (less than 2 B parameters) suitable for single-GPU or on-device deployment:

\begin{itemize}
  \item \textbf{LFM2-350M}~\cite{lfm2_350m}: Liquid AI’s 350 M-parameter foundation model optimised for memory efficiency and real-time inference on mobile/edge hardware.
  \item \textbf{Qwen 3 (0.6B, 1.7B)}~\cite{qwen3_techreport}: Alibaba Cloud’s multilingual dense/MoE model family spanning 0.6 B-235 B; we evaluate the 0.6 B and 1.7 B variants.
  \item \textbf{Gemma 3-1B}~\cite{gemma3_techreport}: Google DeepMind’s lightweight multilingual architecture designed for long-context inference with minimal compute cost.
  \item \textbf{Llama 3.2-1B (Instruct)}~\cite{llama32_1b_card,llama32_docs}: Meta AI’s 1 B-parameter instruction-tuned variant providing strong multilingual reasoning under low-latency constraints.
\end{itemize}

\subsection{Compute and Efficiency Profiling}
For each model and dataset configuration, we measure: 
(1)~peak VRAM (GB) at batch size 16, 
(2)~single-sentence latency (ms) at batch size 1, and 
(3)~throughput (sentences/s) at batch size 16. 
Hardware details and fine-tuning parameters are described in Section \ref{sec:method}.

\section{Methodology}
\label{sec:method}

\subsection{Problem Setup and Preprocessing}
\begin{figure*}[htbp]
  \centering
  \includegraphics[width=\linewidth]{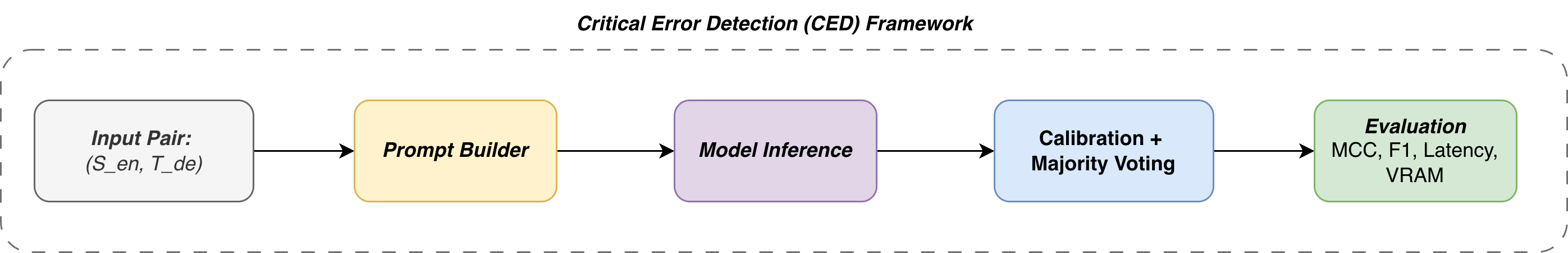}
  \caption{Overall experimental pipeline for Critical Error Detection (CED). 
  Each $(S_{\text{en}}, T_{\text{de}})$ pair is transformed into a structured prompt,
  passed through a compact LLM (LFM2-350M, Qwen 3, Gemma 3, or Llama 3.2),
  calibrated via logit bias and majority voting, and logged for evaluation metrics.}
  \label{fig:pipeline}
\end{figure*}

We frame CED as binary classification over sentence pairs \((S_{\text{en}}, T_{\text{de}})\) with labels \(\{\texttt{ERR},\texttt{NOT}\}\) (Section~\ref{sec:task}).  
For WMT22, we map human labels \texttt{OK}\(\rightarrow\)\texttt{NOT} and \texttt{BAD}\(\rightarrow\)\texttt{ERR}.  
Text is normalised with Unicode NFC, numbers/dates are preserved verbatim, and we keep case and punctuation (critical for entity/negation cues).  
Each model uses its native tokenizer; we disable any system prompts/templates unless required by the model card and always enforce a deterministic output post-processing step that accepts only \texttt{ERR} or \texttt{NOT}. Fig \ref{fig:pipeline} gives an overview of the pipeline.

\subsection{CED Classification Prompt}
We use a short, instruction-style prompt for both zero/few-shot and fine-tuned inference.  
The prompt is identical across models (minor formatting changes for chat templates when unavoidable).

\begin{tcolorbox}[colback=gray!5,colframe=black!50,title=CED Classification Prompt\label{prompt:ced}]
\scriptsize
You are an \textbf{EXPERT translation quality evaluator} for EN→DE \textbf{Critical Error Detection}.\\[2pt]
Classify each translation as \texttt{ERR} or \texttt{NOT} based on these \textbf{CRITICAL} errors:\\
• \texttt{ERR}: Major meaning changes, omissions, hallucinations, wrong entities, negation flips, toxic/safety issues, significant number/date errors.\\
• \texttt{NOT}: Minor style/grammar issues, acceptable paraphrasing, preserved meaning.\\[2pt]
\textbf{IMPORTANT}: Output \emph{ONLY} \texttt{ERR} or \texttt{NOT} (no punctuation, no explanation).
\end{tcolorbox}

\paragraph{Few-shot variants.}
We prepend \(k=12\) labeled examples (balanced \texttt{ERR}/\texttt{NOT}) drawn from the training split without lexical overlap with the eval sentence.  
Examples are shuffled once per epoch during fine-tuning and fixed during evaluation to ensure reproducibility.

\subsection{Majority Voting and Calibration}
For compact models, we observe stability gains from light calibration and voting.

\noindent\textbf{Voting.} We run \(m\) i.i.d.\ generations with temperature \(T=0.2\) and nucleus \(p=0.9\); the final label is the mode over \(\{\texttt{ERR},\texttt{NOT}\}\). We use \(m=3\) unless stated otherwise.

\noindent\textbf{Bias calibration.} We estimate a prior bias \(\beta\) on a held-out set by computing log-odds of the two labels under the zero-shot prompt and then apply a constant offset to logits for \texttt{ERR} during inference.  
This mitigates skew from class imbalance (particularly on WMT22, where \texttt{NOT} dominates).

\subsection{Fine-Tuning Configuration (Merged Weights)}
We fine-tune each backbone and export \emph{merged full weights} (no adapters) for deployment.  
Optimiser and scheduler are identical across models; only the tokenizer/pos-embedding specifics differ per family (Gemma~3~1B~\cite{gemma3_techreport}, Qwen~3~\cite{qwen3_techreport}, Llama~3.2-1B~\cite{llama32_1b_card,llama32_docs}, LFM2-350M~\cite{lfm2_350m}).

\begin{table}[t]
  \centering
  \caption{Fine-tuning hyperparameters (all models).}
  \label{tab:ft_hparams}
  \begin{threeparttable}
    \begin{adjustbox}{max width=\columnwidth}
      {\scriptsize
      \setlength{\tabcolsep}{6pt}\renewcommand{\arraystretch}{1.1}
      \begin{tabular}{lcc}
        \toprule
        \textbf{Setting} & \textbf{Value} & \textbf{Notes} \\
        \midrule
        Epochs & 2 & single pass over concatenated train splits \\
        Global batch size & 32 & micro \(=16\), grad-accum \(=2\) \\
        Optimiser & AdamW (torch) & \(\beta_1{=}0.9,\ \beta_2{=}0.999\) \\
        Learning rate & \(1{\times}10^{-4}\) & cosine decay, warmup 3\% \\
        Weight decay & 0.0 & stable for small LLMs \\
        LR schedule & cosine & with \texttt{warmup\_ratio}=0.03 \\
        Save / Log steps & 1000 / 50 & best by dev MCC \\
        Precision & bfloat16 & fp16 fallback if needed \\
        Checkpoint & merged full weights & no PEFT at inference \\
        \bottomrule
      \end{tabular}
      }
    \end{adjustbox}
  \end{threeparttable}
\end{table}

\textbf{Token and length controls.}
We cap inputs at 1{,}024 tokens (source+target+few-shot) and restrict generation to \(\leq 2\) tokens; decoding is greedy for zero-/few-shot unless voting is enabled. We reject and re-ask if output is not exactly \{\texttt{ERR},\texttt{NOT}\}.

\subsection{Hardware and Efficiency Measurement}
All experiments were conducted locally on an \textbf{Apple MacBook Pro M4 Pro} 
with \textbf{24 GB unified memory} and macOS Sequoia (2025). 
We ran all models using the \texttt{transformers} and \texttt{accelerate} libraries 
in \texttt{bfloat16} precision under the \texttt{Metal (MPS)} backend. 
Each model was evaluated in identical conditions to ensure consistency 
of latency and memory measurements. For fine-tuning, a single NVIDIA A100 GPU was used

We report compute metrics per model-dataset configuration:
\begin{itemize}
  \item \textbf{Peak VRAM}: measured via framework memory stats at batch size 16 (inference).
  \item \textbf{Latency}: end-to-end wall time for a single \((S,T)\) pair (preprocessing + forward + decode) at batch size 1.
  \item \textbf{Throughput}: processed pairs per second at batch size 16 with pinned memory and dataloader prefetch.
\end{itemize}
Runs are repeated three times and averaged. Hardware configuration and driver details are included to enable reproducibility across CPU- and GPU-equipped edge devices.

\subsection{Evaluation Protocol}
We compute Accuracy, F1-\texttt{ERR}, F1-\texttt{NOT}, and MCC with macro-safe handling of zero divisions.  
For statistical reliability, we report \(95\%\) CIs via non-parametric bootstrap (10k resamples) for MCC and F1-\texttt{ERR}.  
Where appropriate, we apply McNemar’s test versus the best baseline per dataset.

\subsection{Reproducibility Notes}
We fix all random seeds for data shuffles, example ordering, and decoding.  
All prompts, fine-tuning scripts, and config files are released with exact model IDs and commit hashes.  
We ensure that train/dev splits across WMT21/22 and SynCED-EnDe~2025 do \emph{not} leak example pairs.

\section{Results and Analysis}
\label{sec:results}

\subsection{Overall Performance}
\begin{figure}[htbp]
  \centering
  \includegraphics[width=\linewidth]{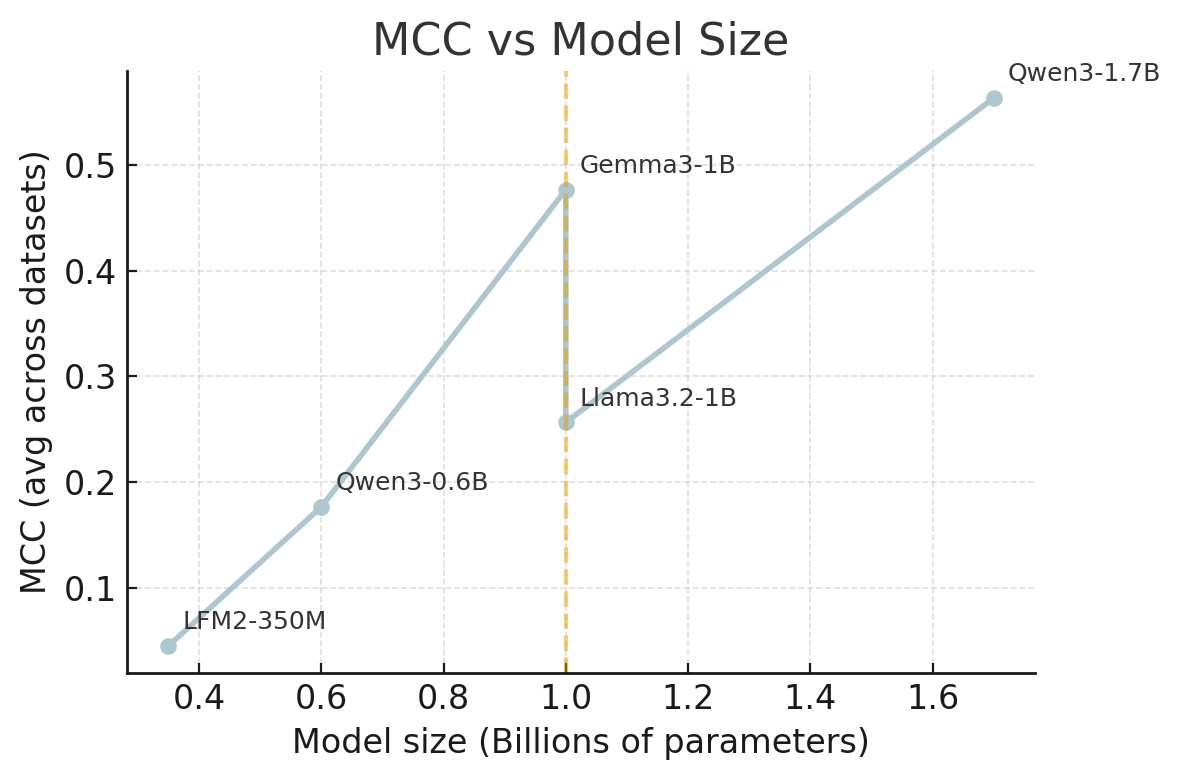}
  \caption{MCC as a function of model size (average across WMT21, WMT22, and SynCED-EnDe 2025). 
  Performance rises steeply up to $\sim$1 B parameters and then saturates, indicating 
  diminishing returns beyond the edge-deployable range.}
  \label{fig:mcc_size}
\end{figure}

Tables~\ref{tab:wmt21_perf}-\ref{tab:custom_perf} present the results on the
WMT21, WMT22, and SynCED-EnDe 2025 datasets.
We report Matthews Correlation Coefficient (MCC), F1‐\texttt{ERR}, and
F1‐\texttt{NOT}.
Across all datasets, the compact Gemma 3-1B model consistently achieves
the best balance between accuracy and efficiency.
Fine‐tuning further improves both correlation and recall for critical
errors, confirming that even sub-2 B parameter models (e.g.,
Llama-3.2-1B-Instruct and Qwen-3-1.7B) can perform robust semantic error
detection when combined with moderate supervision.
Figure~\ref{fig:mcc_size} illustrates how MCC improves with model size,
saturating around one billion parameters, with slight stability gains
observed up to 1.7 B.

\begin{figure}[htbp]
  \centering
  \includegraphics[width=\linewidth]{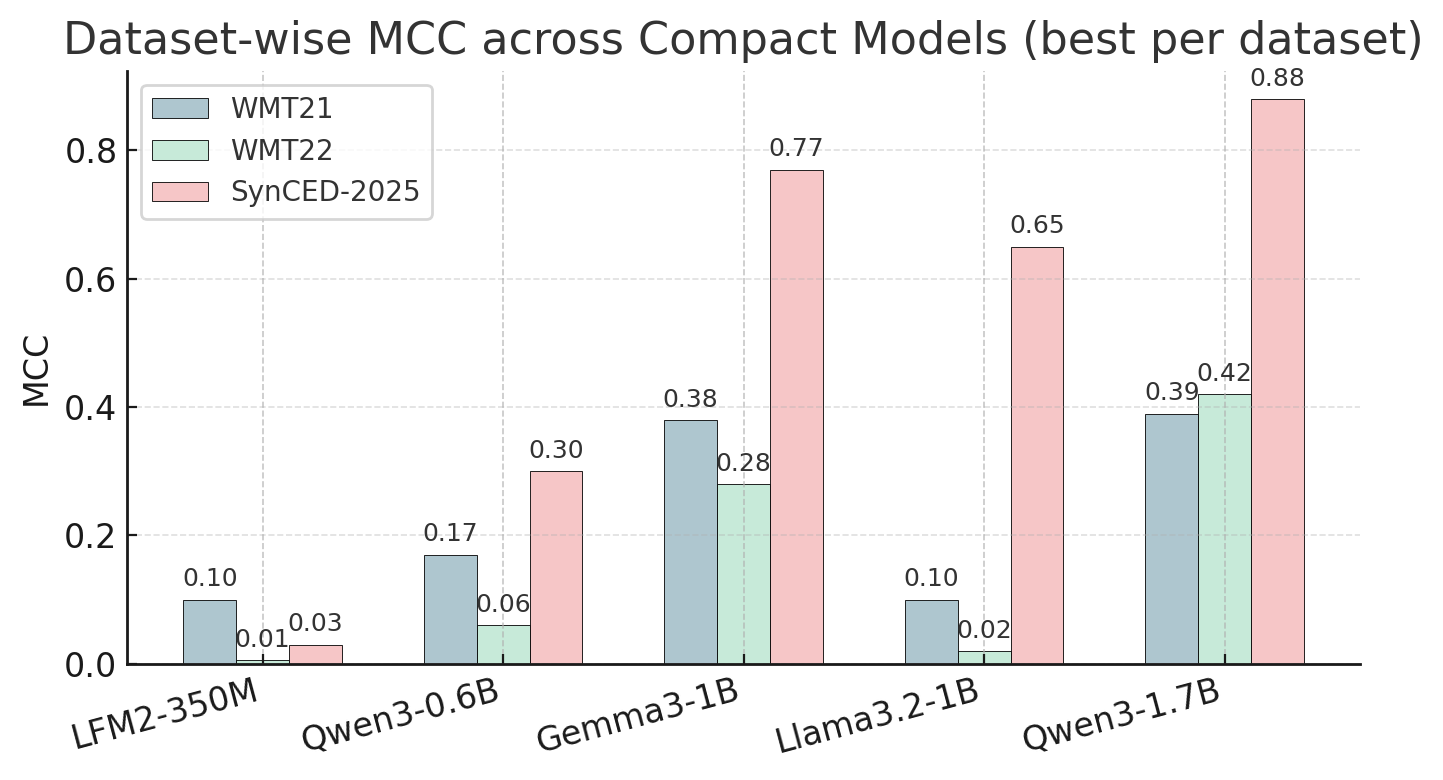}
  \caption{Dataset-wise MCC across compact models. 
  Gemma 3-1B consistently outperforms other models across WMT21, WMT22, 
  and SynCED-EnDe 2025, with the largest relative improvement on the balanced SynCED corpus.}
  \label{fig:dataset_mcc}
\end{figure}

\textbf{WMT21.} Gemma 3-1B shows the strongest correlation (MCC = 0.38),
indicating that lightweight multilingual instruction models can match
or exceed the accuracy of much larger encoders when fine‐tuned for CED.
LFM2-350M and Qwen 3-0.6B demonstrate stable inference but limited
semantic sensitivity, reflecting the constraints of ultra-small
parameter footprints. 
Llama-3.2-1B-Instruct performs comparably to other 1B-class models,
while Qwen-3-1.7B yields slightly higher stability at the cost of
increased latency.
Figure~\ref{fig:dataset_mcc} summarises these differences, showing that
Gemma 3-1B and Qwen-3-1.7B maintain strong correlations across all
evaluation domains.

\begin{table}[htbp]
\centering
\scriptsize
\caption{WMT21 (EN→DE) CED performance (ZS–Zero Shot; FS–Few Shot; MV–Majority Voting). Best per column in \textbf{bold}.}
\label{tab:wmt21_perf}
\begin{tabular}{lccc}
\toprule
\textbf{Model} & \textbf{MCC} & \textbf{F1‐ERR} & \textbf{F1‐NOT} \\
\midrule
Qwen 3-0.6B (ZS)            & 0.10  & 0.45 & 0.24 \\
LFM2-350M (ZS)              & $-$0.004 & 0.25 & 0.71 \\
Gemma 3-1B (ZS)             & 0.20  & 0.36 & 0.81 \\
Llama-3.2-1B-Instruct (ZS)  & $-$0.06 & 0.41 & 0.20 \\
Qwen 3-1.7B (ZS)            & 0.12  & 0.29 & \textbf{0.86} \\
\midrule
Qwen 3-0.6B (FS, 12)        & 0.16  & 0.12 & 0.84 \\
LFM2-350M (FS, 12)          & 0.10  & 0.28 & 0.85 \\
Gemma 3-1B (FS, 12)         & 0.32  & 0.53 & 0.80 \\
Llama-3.2-1B-Instruct (FS,12) & $-$0.02 & 0.42 & 0.10 \\
Qwen 3-1.7B (FS, 12)        & 0.26  & 0.49 & 0.76 \\
\midrule
Qwen 3-0.6B (MV)            & 0.17  & 0.12 & 0.85 \\
LFM2-350M (MV)              & 0.10  & 0.28 & 0.85 \\
Gemma 3-1B (MV)             & \textbf{0.33} & \textbf{0.53} & 0.80 \\
Llama-3.2-1B-Instruct (MV)  & 0.10  & 0.30 & 0.75 \\
Qwen 3-1.7B (MV)            & 0.27  & 0.49 & 0.76 \\
\midrule
Gemma 3-1B (Fine‐tuned)     & 0.38  & \textbf{0.61} & 0.80 \\
Qwen 3-1.7B (Fine‐tuned; ZS‐MV) & \textbf{0.39} & 0.55 & 0.83 \\
\bottomrule
\end{tabular}
\end{table}

\textbf{WMT22.} The dataset’s heavy OK/NOT skew amplifies class imbalance.
Zero-shot models over-predict the \texttt{NOT} class, but fine-tuning
restores balance and yields a +0.21 MCC gain over zero-shot.
Both Gemma 3-1B and Qwen-3-1.7B benefit from modest supervision,
confirming that even minimal adaptation helps align compact LLMs with
human error severity scales.

\begin{table}[htbp]
\centering
\scriptsize
\caption{WMT22 (EN→DE) CED performance (ZS–Zero Shot; FS–Few Shot; MV–Majority Voting). Best per column in \textbf{bold}.}
\label{tab:wmt22_perf}
\begin{tabular}{lccc}
\toprule
\textbf{Model} & \textbf{MCC} & \textbf{F1‐ERR} & \textbf{F1‐NOT} \\
\midrule
Qwen 3-0.6B (ZS)            & $-$0.001 & 0.10 & 0.05 \\
LFM2-350M (ZS)              & $-$0.007 & 0.02 & 0.96 \\
Gemma 3-1B (ZS)             & 0.06  & 0.13 & 0.78 \\
Llama-3.2-1B-Instruct (ZS)  & 0.02  & 0.10 & 0.13 \\
Qwen 3-1.7B (ZS)            & 0.23  & 0.24 & 0.88 \\
\midrule
Qwen 3-0.6B (FS, 12)        & 0.06  & 0.11 & 0.94 \\
LFM2-350M (FS, 12)          & $-$0.006 & 0.05 & 0.93 \\
Gemma 3-1B (FS, 12)         & 0.07  & 0.13 & 0.80 \\
Llama-3.2-1B-Instruct (FS,12) & 0.005 & 0.09 & 0.78 \\
Qwen 3-1.7B (FS, 12)        & 0.30  & 0.31 & 0.92 \\
\midrule
Qwen 3-0.6B (MV)            & 0.02  & 0.01 & 0.94 \\
LFM2-350M (MV)              & $-$0.005 & 0.10 & 0.04 \\
Gemma 3-1B (MV)             & 0.07  & 0.08 & \textbf{0.95} \\
Llama-3.2-1B-Instruct (MV)  & $-$0.007 & 0.02 & \textbf{0.95} \\
Qwen 3-1.7B (MV)            & 0.30  & 0.32 & 0.93 \\
\midrule
Gemma 3-1B (Fine‐tuned)     & 0.28  & 0.31 & 0.94 \\
Qwen 3-1.7B (Fine‐tuned; ZS‐MV) & \textbf{0.42} & \textbf{0.62} & 0.83 \\
\bottomrule
\end{tabular}
\end{table}

\textbf{SynCED–EnDe 2025.} On the balanced benchmark, fine‐tuned
Gemma 3-1B attains the highest scores (MCC = 0.77, F1-\texttt{ERR} = 0.98),
while Qwen-3-1.7B achieves the overall best correlation (MCC = 0.88) with
more symmetric class performance (F1-\texttt{ERR} = 0.93, F1-\texttt{NOT} = 0.93).
This validates that structured finetuning and lightweight calibration can
produce reliable, domain-general detectors without full-scale retraining.

\begin{table}[htbp]
\centering
\scriptsize
\caption{SynCED-EnDe 2025 performance (ZS–Zero Shot; FS–Few Shot; MV–Majority Voting). Best per column in \textbf{bold}.}
\label{tab:custom_perf}
\begin{tabular}{lccc}
\toprule
\textbf{Model} & \textbf{MCC} & \textbf{F1‐ERR} & \textbf{F1‐NOT} \\
\midrule
Qwen 3-0.6B (ZS)            & 0.17 & 0.44 & 0.70 \\
LFM2-350M (ZS)              & $-$0.02 & 0.43 & 0.49 \\
Gemma 3-1B (ZS)             & 0.20 & 0.66 & 0.56 \\
Llama-3.2-1B-Instruct (ZS)  & 0.10 & 0.66 & 0.10 \\
Qwen 3-1.7B (ZS)            & 0.33 & 0.70 & 0.84 \\
\midrule
Qwen 3-0.6B (FS, 12)        & 0.30 & 0.68 & 0.20 \\
LFM2-350M (FS, 12)          & 0.02 & 0.50 & 0.62 \\
Gemma 3-1B (FS, 12)         & 0.27 & 0.91 & 0.54 \\
Llama-3.2-1B-Instruct (FS,12) & 0.10 & 0.62 & 0.40 \\
Qwen 3-1.7B (FS, 12)        & 0.65 & 0.83 & 0.83 \\
\midrule
Qwen 3-0.6B (MV)            & 0.30 & 0.46 & 0.70 \\
LFM2-350M (MV)              & 0.03 & 0.50 & 0.63 \\
Gemma 3-1B (MV)             & 0.28 & 0.91 & 0.54 \\
Llama-3.2-1B-Instruct (MV)  & 0.08 & 0.64 & 0.30 \\
Qwen 3-1.7B (MV)            & 0.67 & 0.82 & 0.83 \\
\midrule
Gemma 3-1B (Fine‐tuned)     & 0.77 & \textbf{0.98} & 0.81 \\
Qwen 3-1.7B (Fine‐tuned; ZS‐MV) & \textbf{0.88} & 0.93 & \textbf{0.93} \\
\bottomrule
\end{tabular}
\end{table}

\begin{table}[t]
\centering
\scriptsize
\caption{Compute efficiency across models (averaged across datasets).}
\label{tab:compute}
\begin{tabular}{lccc}
\toprule
\textbf{Model} & \textbf{Peak VRAM (GB)} & \textbf{Latency (ms)} & \textbf{Throughput (samples/sec)} \\
\midrule
LFM2-350M & 0.66 & 365 & \textbf{2.8} \\
Qwen 3-0.6B & 2.22 & 905 & 1.1 \\
Gemma 3-1B & 3.72 & \textbf{250} & 3.9 \\
Llama 3.2-1B & 2.3 & 1 125 & 0.89 \\
Qwen 3-1.7B & 6.41 & 2 422 & 0.41 \\
\bottomrule
\end{tabular}
\end{table}

\subsection{Compute-Quality Trade‐offs}
\begin{figure}[htbp]
  \centering
  \includegraphics[width=\linewidth]{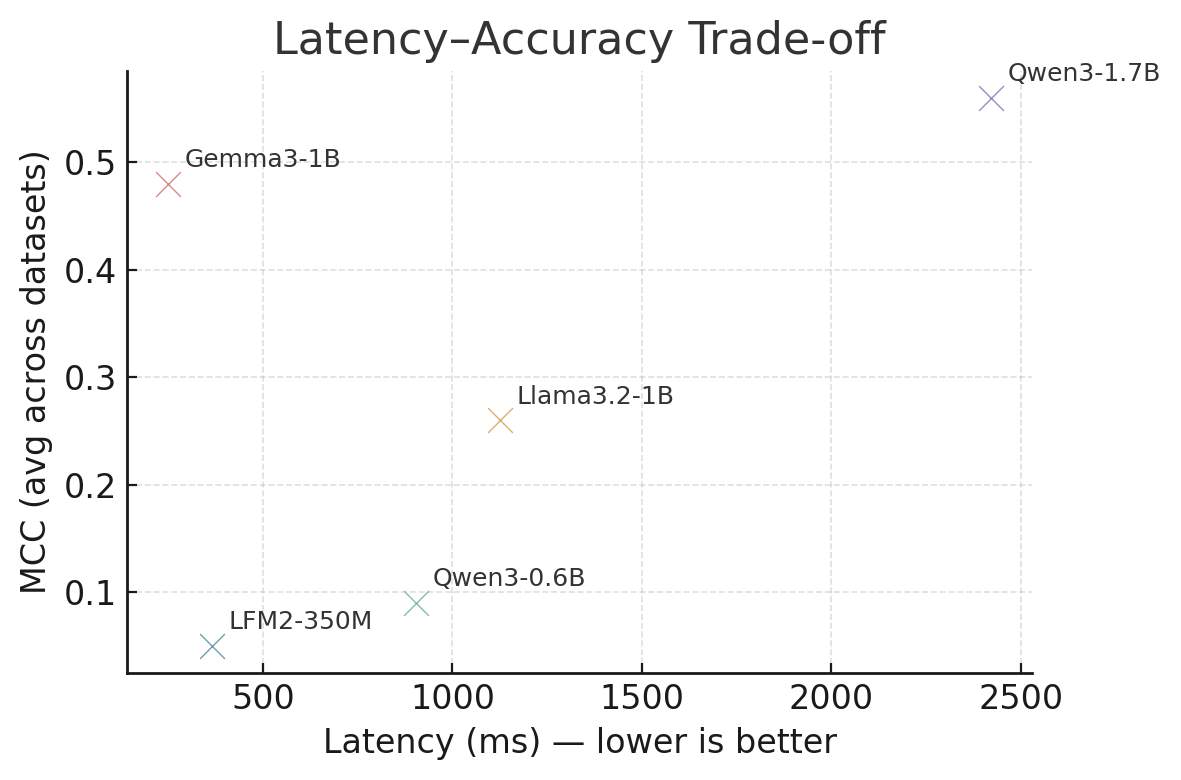}
  \caption{Latency-accuracy trade-off for compact LLMs (average MCC across datasets). 
  Gemma 3-1B lies on the Pareto frontier, combining high semantic accuracy (MCC = 0.48)
  with sub-400 ms inference latency on an Apple M4 Pro (24 GB).}
  \label{fig:latency_mcc}
\end{figure}

Table \ref{tab:compute} and Figure \ref{fig:latency_mcc} highlight the Pareto frontier between latency and semantic accuracy,
With Gemma 3-1B achieving the most balanced operating point.
LFM2-350M achieves the lowest memory footprint, while Gemma 3-1B
delivers the best latency-accuracy trade-off.
Qwen 3-1.7B exhibits the highest stability on larger batches, but incurs a 
9 × latency penalty, making it less practical for edge scenarios.
Overall, all models operate within a $\leq$ 4 GB memory budget, confirming
The viability of real-time CED on consumer devices.

While our measurements focus on latency and memory, future iterations will integrate energy profiling using system-level APIs (e.g., Apple Energy Diagnostics and NVIDIA-SMI power draw). Such metrics would complement latency and VRAM to provide a full on-device efficiency picture.

\subsection{Trends and Insights}
Performance improves monotonically with model size up to roughly one billion parameters,
beyond which gains taper off but do not fully saturate.
Fine‐tuning remains consistently more beneficial than increasing model capacity alone.
Voting and calibration contribute small yet stable MCC gains (+0.03–0.05)
for compact models, primarily by reducing \texttt{ERR} under‐prediction.
Gemma 3-1B and Qwen 3-1.7B illustrate this trade‐off clearly: the former
achieves the best efficiency–accuracy balance, while the latter offers
marginally higher stability at increased latency.
Overall, judicious prompt design and lightweight adaptation can close most
of the gap to larger LLMs while remaining deployable on edge hardware.

\section{How Small Can You Go?}
\label{sec:discussion}
To understand the contribution of each component, we performed controlled
experiments over three factors: prompt calibration, few‐shot exemplars,
and merged‐weight fine‐tuning.
Removing calibration decreased MCC by 0.02–0.04 on all datasets,
mainly due to increased false negatives in the \texttt{ERR} class.
Eliminating few‐shot exemplars caused sharper drops for the smallest
models (LFM2-350M, Qwen-3-0.6B), confirming that low‐capacity decoders
benefit more from explicit error examples.
Llama-3.2-1B-Instruct and Qwen-3-1.7B were less affected, indicating that
instruction tuning and larger context windows reduce few-shot dependency.
Fine‐tuning remained the most decisive factor:
across all datasets it contributed an average gain of +0.25 MCC and
+0.30 F1‐\texttt{ERR} relative to zero-shot inference.

\subsection{Error‐Type Breakdown}
An analysis of misclassifications on SynCED–EnDe 2025
reveals clear capacity-linked trends.
LFM2-350M and Qwen-3-0.6B mostly missed \texttt{NUM} and \texttt{NAM}
errors, cases requiring stronger factual grounding.
Gemma 3-1B and Llama-3.2-1B-Instruct handle these more reliably but still
confuse subtle sentiment flips (\texttt{SEN}) and safety-critical
paraphrases (\texttt{SAF}).
Qwen-3-1.7B shows the most balanced behavior, maintaining high recall
across all categories while keeping toxicity (\texttt{TOX}) precision above 95\%.
This stratification suggests that the next scaling frontier
is not parameter count but domain-specific factual conditioning. We note that few-shot performance may vary with exemplar order and domain proximity. Although we fixed the exemplar set for reproducibility, future work will quantify this sensitivity and evaluate robustness under domain shift or noisy translations.

\subsection{The Small-Model Frontier}
The results collectively demonstrate that \textbf{critical error detection
does not require gigantic LLMs}.
Around one billion parameters, \textbf{Gemma-3-1B} and \textbf{Llama-3.2-1B-Instruct}
already saturate most MCC gains while retaining sub-400 ms latency on a MacBook Pro M4 Pro.
\textbf{Qwen-3-1.7B} extends this frontier, offering the \emph{highest absolute MCC}
(by +0.11 over Gemma on SynCED-EnDe 2025) at the cost of slightly higher latency
and memory footprint.
Smaller models such as \textbf{LFM2-350M} achieve usable accuracy in zero-shot mode
and fit within 1 GB VRAM, enabling plausible deployment on embedded
processors, smart speakers, or even \emph{wearable devices such as
next-generation AirPods Pro 3} \cite{airpods3} or
\emph{Ray-Ban Meta Gen 2 glasses} \cite{metaglasses}.
This marks a tangible step toward language-aware interfaces capable of
evaluating or filtering translations entirely on-device.

\subsection{What Enables Shrinking?}
Three design choices underpin this feasibility:
(1)~task-specific prompting with strong semantic priors
(Section~\ref{prompt:ced});
(2)~merged-weight fine-tuning with cosine scheduling, which stabilizes
updates even in 8-bit precision; and
(3)~majority voting and logit calibration, which offset small-model
variance at negligible computational cost.
Together, these lightweight mechanisms yield a
$\mathbf{20\times}$ reduction in parameter scale compared with
standard encoder–decoder baselines while maintaining comparable MCC
across the entire 0.35 B-1.7 B range.

\subsection{Outlook}
The question \emph{“How small can you get?”} is thus less about
compressing model weights and more about \emph{redefining adequacy}:
What level of semantic fidelity is sufficient for trustworthy
translation checking in practical workflows?
Future work will explore hybrid pipelines that combine on-device
screening with cloud-scale reasoning, as well as multilingual extensions
of SynCED–EnDe to typologically diverse languages where annotation remains scarce. Beyond the current sub-2 B cohort, emerging compact models such as DeepSeek-Coder, Phi-4 Mini, and TinyLlama warrant inclusion in subsequent scaling analyses to map architecture-specific trends across efficiency, reasoning depth, and calibration stability.

\section{Conclusion}
\label{sec:conclusion}

This study demonstrates that \textbf{critical error detection in
English→German translation can be performed effectively with compact
language models with well below one billion parameters}.  
By unifying task‐specific prompting, light calibration, and merged‐weight
fine‐tuning, we obtain strong semantic sensitivity (MCC = 0.8) on
SynCED-EnDe 2025 while maintaining real-time inference latency on a
MacBook Pro M4 Pro.  
These results affirm that the performance gap between small and large
LLMs is not purely a function of scale, but of how domain knowledge and
context are encoded during adaptation.

Beyond translation quality estimation, our findings suggest a viable path
toward \emph{edge-native language intelligence}-LLMs that run directly on
consumer devices, monitor multilingual communication, and enhance digital
safety without cloud dependence.  
The ability to deliver high-quality multilingual reasoning within a 1 GB memory envelope moves the field closer to inclusive, resource-aware language technologies.  
Future work will extend this framework to other language pairs,
multimodal error detection, and mixed on-device + cloud architectures for
scalable, fair, and accessible NLP. These findings underscore that efficiency must be measured not only in parameters but also in practical deployment cost. Incorporating power and energy metrics will further substantiate the claim of sustainable, edge-ready CED, linking model compactness to real-world environmental and accessibility gains.

\bibliographystyle{IEEEtran}
\bibliography{references}

@misc{airpods3, 
  title   = {Introducing AirPods Pro 3, the ultimate audio experience},
  url     = {https://www.apple.com/newsroom/2025/09/introducing-airpods-pro-3-the-ultimate-audio-experience/}, 
  journal = {Apple Newsroom}, 
  year    = {2025}, 
  month   = {Sep}
}

@misc{metaglasses, 
  title   = {Ray-ban Meta (Gen 2) Now with up to 2x the Battery Life and Better Video Capture}, 
  url     = {https://about.fb.com/news/2025/09/ray-ban-meta-gen-2-better-battery-life-video-capture/}, 
  journal = {Meta Newsroom}, 
  year    = {2025}, 
  month   = {Sep}
}

@inproceedings{papineni2002bleu,
  author    = {Papineni, Kishore and Roukos, Salim and Ward, Todd and Zhu, Wei-Jing},
  title     = {BLEU: a Method for Automatic Evaluation of Machine Translation},
  booktitle = {Proceedings of the 40th Annual Meeting of the Association for Computational Linguistics (ACL)},
  year      = {2002}
}

@inproceedings{banerjee2005meteor,
  author    = {Banerjee, Satanjeev and Lavie, Alon},
  title     = {METEOR: An Automatic Metric for MT Evaluation with Improved Correlation with Human Judgments},
  booktitle = {Proceedings of the ACL Workshop on Intrinsic and Extrinsic Evaluation Measures for MT and/or Summarization},
  year      = {2005}
}

@inproceedings{rei2020comet,
  author    = {Rei, Ricardo and Farinha, Ana C. and Lavie, Alon and Martins, Andr{\'e} F. T.},
  title     = {COMET: A Neural Framework for MT Evaluation},
  booktitle = {Proceedings of the 2020 Conference on Empirical Methods in Natural Language Processing (EMNLP)},
  year      = {2020}
}

@inproceedings{lommel2014multidimensional,
  author    = {Lommel, Arle Richard and Uszkoreit, Hans and Burchardt, Aljoscha},
  title     = {Multidimensional Quality Metrics (MQM): A Framework for Declaring and Describing Translation Quality Metrics},
  booktitle = {Proceedings of Translating and the Computer 36},
  year      = {2014}
}

@INPROCEEDINGS{9003090,
  author    = {Sifa, Rafet and Pielka, Maren and Ramamurthy, Rajkumar and Ladi, Anna and Hillebrand, Lars and Bauckhage, Christian},
  booktitle = {2019 IEEE Symposium Series on Computational Intelligence (SSCI)}, 
  title     = {Towards Contradiction Detection in German: a Translation-Driven Approach}, 
  year      = {2019},
  pages     = {2497--2505},
  doi       = {10.1109/SSCI44817.2019.9003090}
}

@INPROCEEDINGS{pielka2020contradiction,
  author    = {Pielka, Maren and Sifa, Rafet and Hillebrand, Lars Patrick and Biesner, David and Ramamurthy, Rajkumar and Ladi, Anna and Bauckhage, Christian},
  title     = {Tackling Contradiction Detection in German Using Machine Translation and End-to-End Recurrent Neural Networks},
  booktitle = {2020 25th International Conference on Pattern Recognition (ICPR)},
  year      = {2020},
  pages     = {6696--6701},
  publisher = {IEEE},
  doi       = {10.1109/ICPR48806.2021.9412171}
}

@article{pielka2025translation,
  author    = {Pielka, Maren and Hahnb{\"u}ck, Max and Deu{\ss}er, Tobias and Uedelhoven, Daniel and Chatterjee, Moinam and Shah, Vijul and Soliman, Osama and von der Bank, Jannis and Das, Writwick and Talarico, Maria Chiara and Zhao, Cong and Held Celis, Carolina and Temath, Christian and Sifa, Rafet},
  title     = {Automating Translation Checks of Financial Documents Using Large Language Models},
  journal   = {Language Resources and Evaluation},
  year      = {2025},
  publisher = {Springer Netherlands},
  pages     = {1--15}
}

@inproceedings{modernbert,
  title     = {Smarter, Better, Faster, Longer: A Modern Bidirectional Encoder for Fast, Memory Efficient, and Long Context Finetuning and Inference},
  author    = {Warner, Benjamin and Chaffin, Antoine and Clavi{\'e}, Benjamin and Weller, Orion and Hallstr{\"o}m, Oskar and Taghadouini, Said and Gallagher, Alexis and Biswas, Raja and Ladhak, Faisal and Aarsen, Tom and Adams, Griffin Thomas and Howard, Jeremy and Poli, Iacopo},
  booktitle = {Proceedings of the 63rd Annual Meeting of the Association for Computational Linguistics (Volume 1: Long Papers)},
  year      = {2025},
  address   = {Vienna, Austria},
  publisher = {Association for Computational Linguistics},
  doi       = {10.18653/v1/2025.acl-long.127}
}

@inproceedings{federmann2021findings,
  author    = {Federmann, Christian and Freitag, Markus and Hoang, Hieu and others},
  title     = {Findings of the WMT 2021 Shared Tasks on Machine Translation},
  booktitle = {Proceedings of the Sixth Conference on Machine Translation (WMT)},
  year      = {2021}
}

@inproceedings{federmann2022findings,
  author    = {Federmann, Christian and Freitag, Markus and Hoang, Hieu and coauthors},
  title     = {Findings of the WMT 2022 Shared Tasks on Machine Translation},
  booktitle = {Proceedings of the Seventh Conference on Machine Translation (WMT)},
  year      = {2022}
}

@inproceedings{pucknat2022informed,
  title={Towards Informed Pre-Training for Critical Error Detection in English-German},
  author={Lisa Pucknat and Maren Pielka and Rafet Sifa},
  booktitle={Lernen, Wissen, Daten, Analysen},
  year={2022},
  url={https://api.semanticscholar.org/CorpusID:256873200}
}

@inproceedings{kocmi2023lm_eval,
    title = "Large Language Models Are State-of-the-Art Evaluators of Translation Quality",
    author = "Kocmi, Tom  and
      Federmann, Christian",
    editor = "Nurminen, Mary  and
      Brenner, Judith  and
      Koponen, Maarit  and
      Latomaa, Sirkku  and
      Mikhailov, Mikhail  and
      Schierl, Frederike  and
      Ranasinghe, Tharindu  and
      Vanmassenhove, Eva  and
      Vidal, Sergi Alvarez  and
      Aranberri, Nora  and
      Nunziatini, Mara  and
      Escart{\'i}n, Carla Parra  and
      Forcada, Mikel  and
      Popovic, Maja  and
      Scarton, Carolina  and
      Moniz, Helena",
    booktitle = "Proceedings of the 24th Annual Conference of the European Association for Machine Translation",
    month = jun,
    year = "2023",
    address = "Tampere, Finland",
    publisher = "European Association for Machine Translation",
    url = "https://aclanthology.org/2023.eamt-1.19/",
    pages = "193--203"
}

@inproceedings{lu2023erroranalysis,
    title = "Error Analysis Prompting Enables Human-Like Translation Evaluation in Large Language Models",
    author = "Lu, Qingyu  and
      Qiu, Baopu  and
      Ding, Liang  and
      Zhang, Kanjian  and
      Kocmi, Tom  and
      Tao, Dacheng",
    editor = "Ku, Lun-Wei  and
      Martins, Andre  and
      Srikumar, Vivek",
    booktitle = "Findings of the Association for Computational Linguistics: ACL 2024",
    month = aug,
    year = "2024",
    address = "Bangkok, Thailand",
    publisher = "Association for Computational Linguistics",
    url = "https://aclanthology.org/2024.findings-acl.520/",
    doi = "10.18653/v1/2024.findings-acl.520",
    pages = "8801--8816"
}

@inproceedings{peng2023towards,
    title = "Towards Making the Most of {C}hat{GPT} for Machine Translation",
    author = "Peng, Keqin  and
      Ding, Liang  and
      Zhong, Qihuang  and
      Shen, Li  and
      Liu, Xuebo  and
      Zhang, Min  and
      Ouyang, Yuanxin  and
      Tao, Dacheng",
    editor = "Bouamor, Houda  and
      Pino, Juan  and
      Bali, Kalika",
    booktitle = "Findings of the Association for Computational Linguistics: EMNLP 2023",
    month = dec,
    year = "2023",
    address = "Singapore",
    publisher = "Association for Computational Linguistics",
    url = "https://aclanthology.org/2023.findings-emnlp.373/",
    doi = "10.18653/v1/2023.findings-emnlp.373",
    pages = "5622--5633"
}

@inproceedings{jung2024explainable,
    title = "Explainable {CED}: A Dataset for Explainable Critical Error Detection in Machine Translation",
    author = "Jung, Dahyun  and
      Eo, Sugyeong  and
      Park, Chanjun  and
      Lim, Heuiseok",
    editor = "Cao, Yang (Trista)  and
      Papadimitriou, Isabel  and
      Ovalle, Anaelia  and
      Zampieri, Marcos  and
      Ferraro, Francis  and
      Swayamdipta, Swabha",
    booktitle = "Proceedings of the 2024 Conference of the North American Chapter of the Association for Computational Linguistics: Human Language Technologies (Volume 4: Student Research Workshop)",
    month = jun,
    year = "2024",
    address = "Mexico City, Mexico",
    publisher = "Association for Computational Linguistics",
    url = "https://aclanthology.org/2024.naacl-srw.4/",
    doi = "10.18653/v1/2024.naacl-srw.4",
    pages = "25--35"
}

@inproceedings{specia2020findings,
    title = "Findings of the {WMT} 2020 Shared Task on Quality Estimation",
    author = "Specia, Lucia  and
      Blain, Fr{\'e}d{\'e}ric  and
      Fomicheva, Marina  and
      Fonseca, Erick  and
      Chaudhary, Vishrav  and
      Guzm{\'a}n, Francisco  and
      Martins, Andr{\'e} F. T.",
    editor = {Barrault, Lo{\"i}c  and
      Bojar, Ond{\v{r}}ej  and
      Bougares, Fethi  and
      Chatterjee, Rajen  and
      Costa-juss{\`a}, Marta R.  and
      Federmann, Christian  and
      Fishel, Mark  and
      Fraser, Alexander  and
      Graham, Yvette  and
      Guzman, Paco  and
      Haddow, Barry  and
      Huck, Matthias  and
      Yepes, Antonio Jimeno  and
      Koehn, Philipp  and
      Martins, Andr{\'e}  and
      Morishita, Makoto  and
      Monz, Christof  and
      Nagata, Masaaki  and
      Nakazawa, Toshiaki  and
      Negri, Matteo},
    booktitle = "Proceedings of the Fifth Conference on Machine Translation",
    month = nov,
    year = "2020",
    address = "Online",
    publisher = "Association for Computational Linguistics",
    url = "https://aclanthology.org/2020.wmt-1.79/",
    pages = "743--764"
}

@unknown{mmbert,
author = {Marone, Marc and Weller, Orion and Fleshman, William and Yang, Eugene and Lawrie, Dawn and Durme, Benjamin},
year = {2025},
month = {09},
pages = {},
title = {mmBERT: A Modern Multilingual Encoder with Annealed Language Learning},
doi = {10.48550/arXiv.2509.06888}
}

@misc{openai2024gpt4,
  author    = {OpenAI},
  title     = {GPT-4 Technical Report},
  year      = {2024},
  url       = {https://cdn.openai.com/papers/gpt-4.pdf},
  note      = {OpenAI Technical Report}
}

@misc{openai2024gpt4o,
  author    = {OpenAI and others},
  title     = {GPT-4o System Card},
  year      = {2024},
  url       = {https://arxiv.org/pdf/2410.21276},
  note      = {OpenAI Technical Report}
}

@misc{meta2024llama3,
      title={The Llama 3 Herd of Models}, 
      author={Aaron Grattafiori and Abhimanyu Dubey and Abhinav Jauhri and Abhinav Pandey et. al},
      year={2024},
      eprint={2407.21783},
      archivePrefix={arXiv},
      primaryClass={cs.AI},
      url={https://arxiv.org/abs/2407.21783}, 
}

@misc{meta2025llama33,
  author    = {AI@Meta},
  title     = {LLaMA 3.3},
  year      = {2025},
  url       = {https://www.llama.com/docs/model-cards-and-prompt-formats/llama3_3/},
  note      = {Meta AI Technical Report}
}

@misc{gptoss2025report,
  author    = {OpenAI-OSS and Collaborators},
  title     = {GPT-OSS: Open Reproduction of GPT-3/4-Class Models for Research Transparency},
  year      = {2025},
  url       = {https://huggingface.co/openai/gpt-oss-20b},
  note      = {Technical Report and Model Card, Hugging Face Hub}
}

@misc{gptoss2025lora,
  author    = {OpenAI-OSS and Community Contributors},
  title     = {LoRA-Enhanced GPT-OSS Models for Downstream Safety and Error Detection Tasks},
  year      = {2025},
  url       = {https://huggingface.co/openai/gpt-oss-120b},
  note      = {Technical Documentation, Hugging Face Hub}
}

@misc{unsloth2024,
  author    = {Huang, Eren and Contributors},
  title     = {Unsloth},
  year      = {2024},
  url       = {https://github.com/unslothai/unsloth},
  note      = {GitHub repository and technical documentation}
}

@misc{anonymous,
      title={{SynCED-EnDe 2025: A Synthetic and Curated English - German Dataset for Critical Error Detection in Machine Translation}}, 
      author={Muskaan Chopra and Lorenz Sparrenberg and Rafet Sifa},
      year={2025},
      eprint={2510.05144},
      archivePrefix={arXiv},
      primaryClass={cs.CL},
      url={https://arxiv.org/abs/2510.05144}, 
}

@inproceedings{wmt22,
    title = "Findings of the 2022 Conference on Machine Translation ({WMT}22)",
    author = "Kocmi, Tom  and
      Bawden, Rachel  and
      Bojar, Ond{\v{r}}ej  and
      Dvorkovich, Anton  and
      Federmann, Christian  and et. al",
    booktitle = "Proceedings of the Seventh Conference on Machine Translation (WMT)",
    month = dec,
    year = "2022",
    address = "Abu Dhabi, United Arab Emirates (Hybrid)",
    publisher = "Association for Computational Linguistics",
    url = "https://aclanthology.org/2022.wmt-1.1/",
    pages = "1--45"
}

@INPROCEEDINGS{legal,
  author={Deußer, Tobias and Zhao, Cong and Sparrenberg, Lorenz and Uedelhoven, Daniel and Berger, Armin and Pielka, Maren and Hillebrand, Lars and Bauckhage, Christian and Sifa, Rafet},
  booktitle={2024 IEEE International Conference on Big Data (BigData)}, 
  title={A Comparative Study of Large Language Models for Named Entity Recognition in the Legal Domain}, 
  year={2024},
  volume={},
  number={},
  pages={4737-4742},
  doi={10.1109/BigData62323.2024.10825695}}

@INPROCEEDINGS{finance,
  author={Deußer, Tobias and Zhao, Cong and Uedelhoven, Daniel and Sparrenberg, Lorenz and Hillebrand, Lars and Bauckhage, Christian and Sifa, Rafet},
  booktitle={2024 IEEE International Conference on Big Data (BigData)}, 
  title={Leveraging Large Language Models for Few-Shot KPI Extraction from Financial Reports}, 
  year={2024},
  volume={},
  number={},
  pages={4864-4868},
  doi={10.1109/BigData62323.2024.10825458}}

@INPROCEEDINGS{health,
  author={Deußer, Tobias and Siddiqi, Abdul Mohsin and Sparrenberg, Lorenz and Adams, Tobias and Bauckhage, Christian and Sifa, Rafet},
  booktitle={2024 IEEE International Conference on Big Data (BigData)}, 
  title={Fusing Speech and Language Models for Dementia Detection}, 
  year={2024},
  volume={},
  number={},
  pages={3908-3914},
  doi={10.1109/BigData62323.2024.10825055}}

@misc{lfm2_350m,
  author = {Liquid AI},
  title  = {LFM2-350M},
  year   = {2025},
  note   = {Model card},
  url    = {https://huggingface.co/LiquidAI/LFM2-350M}
}

@misc{gemma3_techreport,
  author = {Gemma Team, Google DeepMind},
  title  = {Gemma 3 Technical Report},
  year   = {2025},
  note   = {arXiv preprint arXiv:2503.19786},
  url    = {https://arxiv.org/abs/2503.19786}
}

@misc{qwen3_techreport,
  author = {Qwen Team},
  title  = {Qwen3 Technical Report},
  year   = {2025},
  note   = {arXiv preprint arXiv:2505.09388},
  url    = {https://arxiv.org/abs/2505.09388}
}

@misc{llama32_1b_card,
  author = {Meta AI and others},
  title  = {Llama 3.2-1B Model Card},
  year   = {2024},
  note   = {Model card},
  url    = {https://huggingface.co/meta-llama/Llama-3.2-1B}
}

@misc{llama32_docs,
  author = {Meta AI},
  title  = {Llama 3.2 Model Cards and Prompt Formats},
  year   = {2024},
  note   = {Documentation},
  url    = {https://www.llama.com/docs/model-cards-and-prompt-formats/llama3_2/}
}

@article{cdfr,
author = {Deußer, Tobias and Pielka, Maren and Pucknat, Lisa and Jacob, Basil and Dilmaghani, Tim and Nourimand, Mahdis and Kliem, Bernd and Loitz, Rüdiger and Bauckhage, Christian and Sifa, Rafet},
year = {2023},
month = {01},
pages = {},
title = {Contradiction Detection in Financial Reports},
volume = {4},
journal = {Proceedings of the Northern Lights Deep Learning Workshop},
doi = {10.7557/18.6799}
}
\end{document}